\documentclass{svproc}
\usepackage{url}
\usepackage{graphicx}

\begin{document}
	\mainmatter              
	\title{Automating Document Intelligence in Statutory City Planning}
	\titlerunning{Automating Document Intelligence}  
	%
	\author{Lars Malmqvist\inst{1} \and Robin Barber\inst{2}}
	\authorrunning{Malmqvist and Barber} 
	\institute{Research and Implementation, Copenhagen, Denmark\\
		\email{lars@resai.dk}
		\and
		Arcus Global, Cambridge, UK\\
		\email{robin.barber@arcusglobal.com}}
	
	\maketitle              
	
	\begin{abstract}
		UK planning authorities face a legislative conflict between the Planning Act, which mandates public access to application documents, and the Data Protection Act, which requires protection of personal information. This situation creates a manually intensive workload for processing large document volumes, diverting planning officers to administrative tasks and creating legal compliance risks. This paper presents an integrated AI system designed to address these challenges. The system automates the identification and redaction of personal information, extracts key metadata from planning documents, and analyzes architectural drawings for specified features. It operates with an \emph{AI-in-the-Loop (AI2L)} design, presenting all suggestions for review and confirmation by planning officers directly within their existing software; no action is committed without explicit human approval. The system is designed to improve its performance over time by learning from this human oversight through active learning prioritization rather than auto-approval. The system is currently being piloted at four diverse UK local authorities. The paper details the system design, the AI2L workflow, and the evaluation framework used in the pilot. Additionally, it describes a preliminary Return on Investment (ROI) model developed to quantify potential savings and secure partner participation. This work provides a case study on deploying AI to reduce administrative burden and manage compliance risk in a public sector environment.
	\end{abstract}
	\section{Introduction}
	
	The administration of public services, particularly in urban planning, is a document-driven process. Planning departments manage large volumes of semi-structured and unstructured documents, including applications, architectural drawings, zoning codes, and public feedback \cite{Kitchin2016}. This reliance on manual processing creates systemic inefficiencies, diverting skilled planners from high-value strategic work to repetitive administrative tasks. The challenge is not merely one of volume but of complexity, as planners must synthesize information from multimodal sources to make decisions \cite{Kitchin2016}.
	
	A significant driver for technological intervention in the UK is the legal conflict between the Town and Country Planning Act, which mandates public access to planning documents, and the Data Protection Act (2018), which requires stringent protection of personal information. This situation creates a high-risk, labor-intensive need for document redaction, adding to the administrative burden and creating clear compliance risks. These issues are representative of the legal, regulatory, and data-related challenges common in public sector AI adoption \cite{Zuiderwijk2021}.
	
	Intelligent Document Processing (IDP) offers a technological response, moving beyond simple Optical Character Recognition (OCR) to a more advanced understanding of document text, layout, and imagery. A key development in this area was the introduction of models that jointly pre-train textual and layout information, establishing a new approach for document understanding tasks \cite{Xu2020}.
	
	This paper reports on an AI system developed from a Proof-of-Concept to a multi-authority pilot, designed to address the challenges in planning administration. The system provides a case study of implementing an ``AI-in-the-Loop'' (AI2L) system. This is a collaborative framework where the human expert remains the primary decision-maker, using AI as a sophisticated assistant. This approach contrasts with a ``Human-in-the-Loop'' model, where the AI is the primary actor that escalates to a human for exceptions \cite{Eiband2021}. The AI2L framework was selected to navigate the sociotechnical challenges inherent in public sector AI deployment, ensuring that human accountability is maintained in a high-stakes environment \cite{Zuiderwijk2021}.
	
	The contributions of this paper are:
	\begin{itemize}
		\item A case study of an applied AI system addressing a legally-mandated, high-risk administrative task in a government setting.
		\item A description of the system's development from a rapid PoC to an integrated, multi-authority pilot.
		\item A detailed account of the system's AI2L workflows, designed to support rather than replace human expertise.
		\item A report on a preliminary Return on Investment (ROI) model used as a practical tool to facilitate adoption in the public sector.
	\end{itemize}
	
	\section{Related Work}
	
	This work is situated at the intersection of three domains: technical research in Intelligent Document Processing, UK public sector digitisation initiatives, and legal and governance frameworks for AI in government.
	
	\subsection{Intelligent Document Processing}
	Research on intelligent document processing (IDP) has moved beyond simple text recognition to models that jointly reason over text, layout, and visual structure \cite{abdallah2024_form_survey}. Pre-training with layout-aware objectives has been influential; \textit{LayoutLM} demonstrated that incorporating 2D token coordinates improves performance on a range of document understanding tasks \cite{Xu2020}. Progress in the field has been supported by the development of public datasets for layout detection and form understanding \cite{zhong2019_publaynet, jaume2019_funsd} and by the emergence of tooling ecosystems like \textit{LayoutParser}, which provides reusable components for deploying document analysis pipelines \cite{shen2021_layoutparser}. Our work applies these foundational IDP techniques to a specific administrative workflow.
	
	\subsection{Digitisation in UK Planning}
	In the UK, government-led programmes have sought to replace document-centric planning processes with structured, machine-readable data. The Open Digital Planning (ODP) partnership coordinates the development of products such as the Back Office Planning System (BOPS) across multiple local authorities \cite{bops_localdigital}. Complementary initiatives target legacy records; the \textit{Extract} project, a collaboration between the AI industry and the Department for Levelling Up, Housing and Communities (DLUHC), applies AI to convert historic maps and PDFs into planning constraints data \cite{govuk_press_release_2025}. The national Planning Data Platform provides a harmonised backbone for publishing these structured datasets \cite{planning_data_platform}. Our system complements these efforts by focusing on the point of document ingestion, creating structured data from new applications before they enter back-office systems.
	
	\subsection{Legal and Information Governance}
	The deployment of IDP in UK planning is shaped by specific legal and governance constraints. Article 40 of the Town and Country Planning (Development Management Procedure) (England) Order 2015 requires authorities to maintain a public planning register, including online publication of documents \cite{dmpeo2015_article40}. Concurrently, the UK GDPR and Data Protection Act 2018 impose duties to minimise and protect personal data \cite{lga2021_planning_gdpr}. Guidance from the UK's Information Commissioner’s Office (ICO) sets out secure practices for disclosing documents, including the need for robust redaction \cite{ico2025_disclosing}. Research into PDF security highlights the risks of improper redaction, where residual information in the file can allow adversaries to recover masked text, underscoring the need for safe redaction workflows \cite{bland2023_deredaction}. The system is designed to operate within this framework, providing tools to assist with these redaction and disclosure obligations.
	
	\subsection{Human-AI Collaboration Frameworks}
	The design of the system is informed by research in Human-Computer Interaction (HCI) on human-AI collaboration. Guidelines for systems that advise experts recommend keeping the human in control and making the AI's capabilities and limitations clear \cite{amershi2019_guidelines}. Empirical studies show that the most accurate autonomous model is not always the best collaborator; predictable models that enable effective human oversight can yield higher human-AI team performance \cite{bansal2021_teammate}. Our choice of an AI-in-the-Loop (AI2L) pattern aligns with this research. It also reflects UK government principles for AI governance, which emphasize transparency, accountability, and risk-based oversight for algorithmic tools used in public service delivery \cite{atrs2025, uk_whitepaper_2023}.
	
	\section{System Architecture and Implementation}
	
	The system was developed through a structured, multi-phase lifecycle designed to de-risk technology adoption and align the final product with user needs. This section details the system's progression from an initial Proof of Concept (PoC) to the current integrated pilot, and outlines the planned future phases based on pilot feedback.
	
	\subsection{Phase 1: Proof of Concept}
	The project began with a rapid PoC to test the core hypothesis: that contemporary Vision-Language Models (VLMs) could perform essential document understanding tasks on planning documents with sufficient accuracy. The PoC system was a standalone Python application with a FastAPI interface, allowing for quick iteration on the VLM interaction logic. A lightweight demonstration interface using Streamlit enabled early, frequent feedback from project partners. The primary goal of this phase was to establish technical feasibility for three key use cases: information extraction, PII redaction, and basic planning validation checks. The success of the PoC, which confirmed the viability of a VLM-based approach, provided the justification to proceed to a full pilot.
	
	\subsection{Phase 2: Integrated Pilot System}
	The pilot phase focused on transforming the PoC's validated concepts into a robust, secure, and usable application integrated directly into the planners' daily workflow. This required a significant evolution of the system architecture.
	
	\subsubsection{Enterprise Integration and Security}
\begin{figure}[t]
		\centering
		\includegraphics[width=0.8\linewidth]{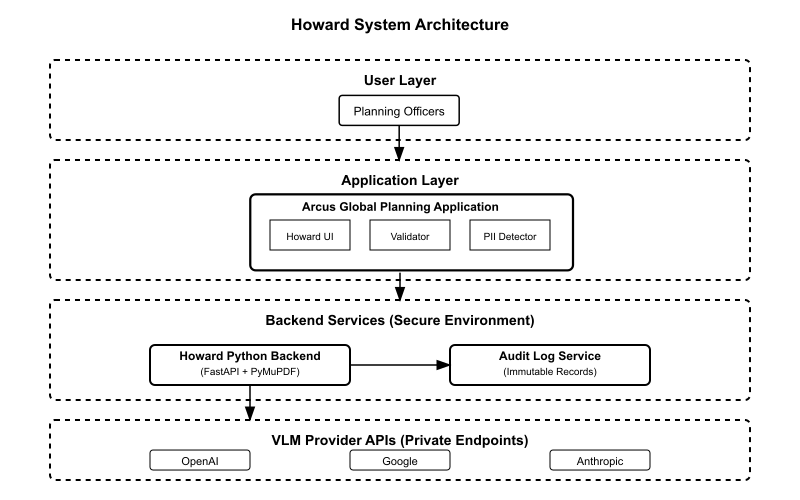} 
		\caption{The pilot system architecture, highlighting the integration of the system backend with the existing enterprise application and its secure, auditable interaction with VLM services.}
		\label{fig:architecture}
	\end{figure}

	The standalone architecture was replaced with an integrated, enterprise-grade design. The core Python backend is now deployed in a secure environment, communicating with VLM provider APIs (e.g., OpenAI, Google, Anthropic) exclusively over private network endpoints. The Streamlit demo UI was retired in favor of custom components built directly into the partner's enterprise planning application. This deep integration, shown in Figure \ref{fig:architecture}, is a central design feature, ensuring that the system's capabilities are presented as a natural extension of the user's existing tools. To address reviewer concerns about figure legibility, all figures in this version have been re-rendered at full column width with enlarged labels; where possible, vector assets are used for clarity.
	
	\subsubsection{Configurable Hybrid Processing Framework}
	The pilot system introduced a configurable framework to manage the interaction with VLMs, designed to ensure reliability and auditability. Instead of hard-coded logic, the system uses sequences of configurable actions for pre-processing and post-processing.
	\begin{itemize}
		\item \textbf{Pre-processing:} Before a request is sent to a VLM, a series of pre-processing steps are executed. These include rendering PDF pages to images using the PyMuPDF library, using classic computer vision techniques to focus the VLM on specific regions of interest, and dynamically constructing detailed prompts from templates.
		\item \textbf{Post-processing:} After the VLM responds, a corresponding sequence of post-processing actions is triggered. This includes validating the VLM's output against an expected data schema, applying rule-based heuristics to filter out potential false positives, and normalizing the data into a standard format.
		\item \textbf{Error Correction and Auditability:} The pilot framework includes automated retry logic. If a VLM call fails or the response is invalid, the system can make a second attempt using a more explicit "fallback" prompt. Every action taken by the framework, from the initial request to the final validated output, including any retry attempts, is recorded in an immutable audit log.
	\end{itemize}
	
	\paragraph{Division of Responsibilities and Scope.}
	Consistent with the AI2L paradigm, the system proposes and the planner disposes: suggested metadata values are verified and corrected by the officer prior to commit; candidate PII items are presented for confirmation before any redaction is applied; and visual detections are provided as overlays to support, not supplant, human judgment. We do not expose a general-purpose Q\&A chatbot in the pilot. Instead, VLMs are used for structured extraction, detection, and layout-aware analysis whose outputs are then validated and, where relevant, evaluated by deterministic checks prior to human confirmation.
	
	\section{The AI-in-the-Loop Workflow}
	
	In high-stakes public sector domains like urban planning, where decisions have legal and social consequences, the objective of AI implementation is the augmentation of professional judgment, not its replacement. This necessitates a system design that maintains human accountability. Research in Human-Computer Interaction (HCI) distinguishes between two models of collaboration. In a Human-in-the-Loop (HITL) model, the AI performs a task autonomously and escalates to a human for exceptions or low-confidence cases. In an AI-in-the-Loop (AI2L) model, the human expert remains the primary actor, using the AI as an assistive tool to support their analysis and decision-making \cite{Eiband2021}.
	
	The system is explicitly designed using the AI2L paradigm. This choice aligns with established guidelines for human-AI interaction, which recommend that systems clarify their capabilities and allow for user correction while keeping the human in control of the process \cite{amershi2019_guidelines}. The goal is to create an effective human-AI team, where the AI functions as a predictable and reliable collaborator \cite{bansal2021_teammate}. The following subsections detail the three primary AI2L workflows implemented in the system.
	
	\subsection{Data Extraction and Verification}
\begin{figure}[t]
		\centering
		\includegraphics[width=0.8\linewidth]{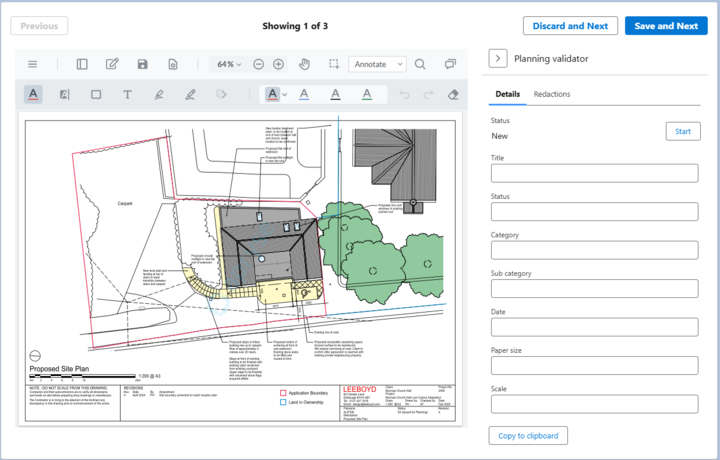}
		\caption{The data extraction workflow. The system populates fields on the right-hand panel based on its analysis of the document shown on the left. The human operator verifies and confirms this data.}
		\label{fig:validator_ui}
	\end{figure}

	The first workflow assists planners with populating metadata fields for new documents. After a document is ingested, the system's data extraction service analyzes its content and automatically populates the corresponding fields in the user interface, such as `Title`, `Date`, and `Scale`. As shown in Figure \ref{fig:validator_ui}, these AI-generated suggestions are presented to the planner for review. The planner's task is to verify the accuracy of each field, make any necessary corrections, and provide final confirmation by saving the record. The human operator retains full authority over the data that is entered into the system of record.
	
	\subsection{Review of Detected Personal Information}
\begin{figure}[t]
		\centering
		\includegraphics[width=0.8\linewidth]{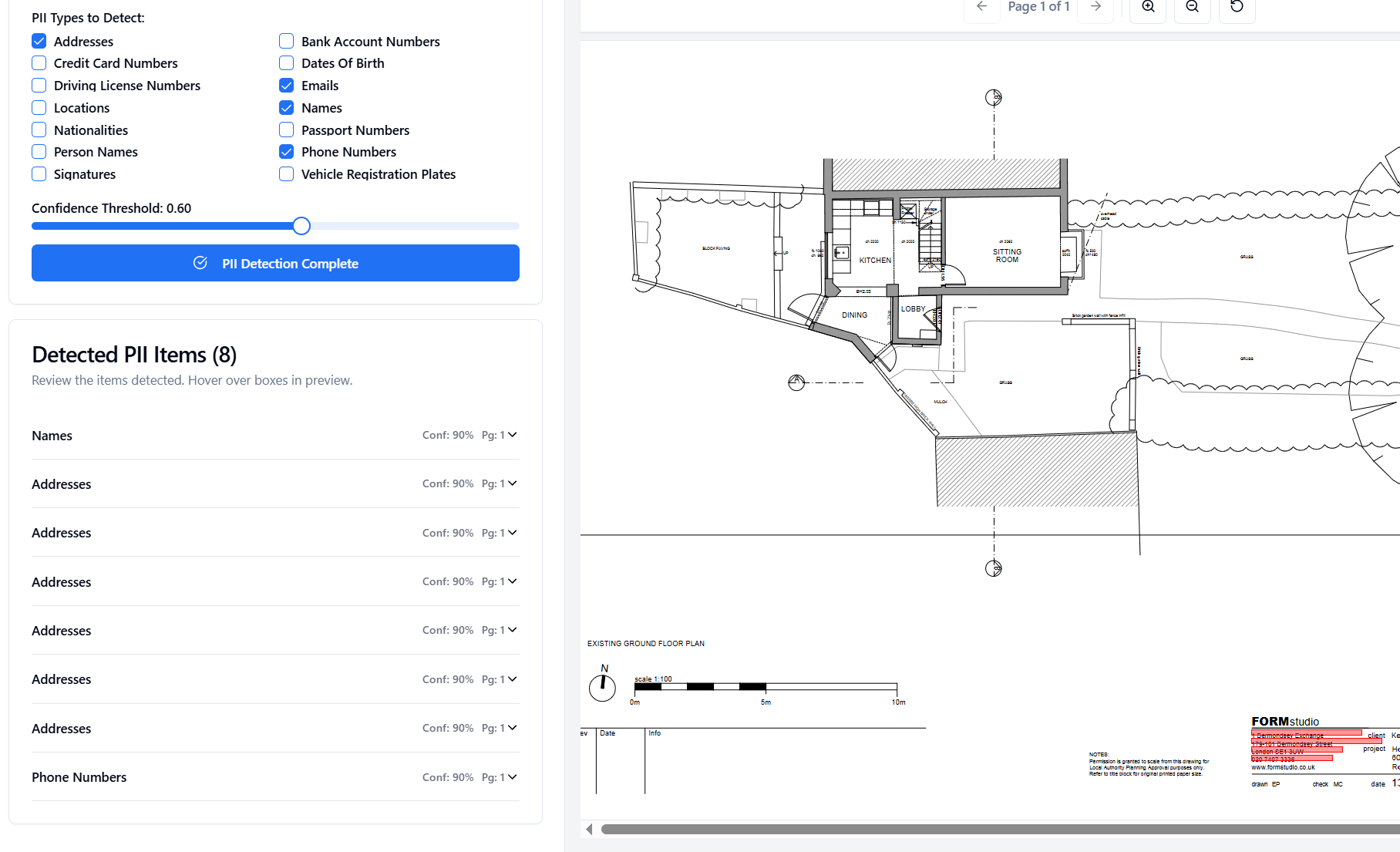}
		\caption{The PII detection workflow. The system presents a list of detected PII items with confidence scores. The operator reviews the list and confirms which items to redact.}
		\label{fig:pii_ui}
	\end{figure}

	The second workflow supports compliance with data protection regulations. The PII detection service analyzes the document and generates a list of all potential instances of personally identifiable information. This list is presented to the user for review, as shown in Figure \ref{fig:pii_ui}. Each detected item is presented with its type (e.g., `Names`, `Emails`), its value, and a VLM-generated confidence score. The planner reviews this list and provides confirmation for each item that should be redacted. The system does not perform any redaction without explicit human instruction for each detected entity. This workflow keeps the user in control of the sensitive redaction process.
	
	\subsection{Configuration and Review of Visual Checks}
\begin{figure}[t]
		\centering
		\includegraphics[width=0.8\linewidth]{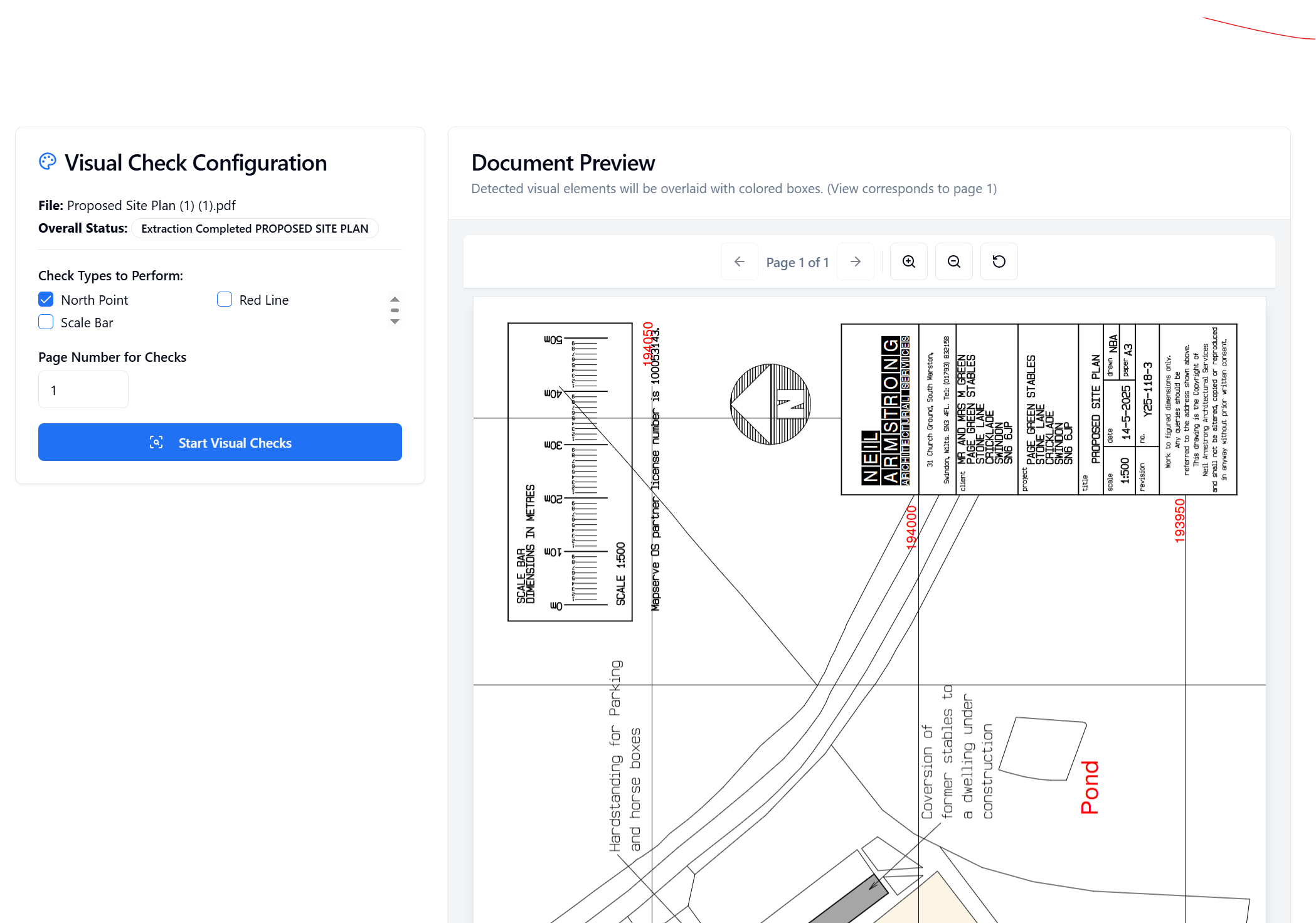}
		\caption{The visual check workflow. The operator selects the checks to perform (left panel), and the system returns a document preview with the detected visual elements overlaid with bounding boxes (right panel).}
		\label{fig:visual_check_ui}
	\end{figure}

	The third workflow allows planners to perform targeted visual analysis of site plans. The user first configures the task by selecting the specific elements they wish to locate from a predefined list, such as a North Point or a Scale Bar (Figure \ref{fig:visual_check_ui}, left). The system then analyzes the document and overlays bounding boxes on the document preview for any detected elements (Figure \ref{fig:visual_check_ui}, right). The final output is a visual aid for the planner, who uses this information to support their own assessment of the plan's validity. The human operator initiates the check and interprets the visual output.
	
	\paragraph{Example: A Deterministic Validation Rule.}
	To make the visual-check workflow concrete, consider a commonly applied requirement for a Site Location Plan: the presence of (i) a North arrow and (ii) a valid scale indicator (e.g., 1:1250 or 1:2500) on the sheet. In the system, symbol detection proposes candidate regions for a North arrow and a scale bar, and text extraction operates over regions of interest where scale text is expected. A deterministic evaluator then checks that at least one North arrow has been detected with sufficient confidence and that extracted scale text matches a jurisdiction-configured regular expression for acceptable scales. The interface overlays these findings on the document and invites the officer to confirm or override. This pattern illustrates the broader design: structured proposals from a VLM are verified by simple, explicit rules and are always subject to human confirmation.
	
	Taken together, these three workflows demonstrate tailored applications of the AI-in-the-Loop principle, corresponding to different user tasks: verifying populated data fields, confirming items in a generated list, and interpreting visual aids. In each case, the system presents AI-generated information as a suggestion, and the final judgment, action, and legal accountability remain with the human operator. This design approach directly addresses the requirements for meaningful human control in high-stakes public sector applications and serves as the foundation for the system's evaluation in the pilot study.
	
	\section{Deployment and Evaluation}
	
	The deployment of the system followed a phased, pilot-based methodology, a recommended practice for managing risk and demonstrating value in public sector AI implementations \cite{Zuiderwijk2021}. This section details the pilot study's design, the framework used for evaluation, and the role of a preliminary Return on Investment (ROI) model in facilitating the project.
	
	\subsection{Pilot Design and Timeline}
	Following the successful completion of the Proof of Concept phase (April - July 2025), the integrated pilot system was developed for deployment. The pilot commenced in Fall 2025 at four local planning authorities in the UK, selected to represent a range of planning environments, as detailed in Table \ref{tab:pilot_authorities}. This diversity is intended to test the system's applicability across different operational contexts, from dense urban areas with major development projects to smaller, primarily residential or rural areas. The pilot involves a cohort of planning officers and administrative staff from each authority who were trained to use the system for their daily document processing tasks.
	
	\begin{table}
		\centering
		\caption{Profiles of the Four UK Local Planning Authorities Participating in the Pilot Study}
		\label{tab:pilot_authorities}
		\begin{tabular}{lllc}
			\hline
			\textbf{Authority} & \textbf{Type} & \textbf{Key Characteristics} & \textbf{Approx. Apps./Year} \\
			\hline
			Authority A & Large Unitary & Primarily rural, large geographic area & 6,000 \\
			Authority B & Unitary & High-growth urban area & 4,500 \\
			Authority C & London Borough & Dense urban, mixed small and major projects & 5,000 \\
			Authority D & Unitary & Industrial and residential focus & 2,000 \\
			\hline
		\end{tabular}
	\end{table}
	
	\subsection{Evaluation Framework}
	\label{sec:evaluation}
	
	This section documents datasets, protocols, metrics, cost, and comparisons. \emph{For clarity: all quantitative results reported in this section derive from the PoC conducted in a controlled setting and do not reflect live usage in the pilot authorities.} Real-world data is being collected as part of the Pilot.
	
	\subsubsection*{Dataset and Annotation Protocol}
	
	We curated a corpus of \textbf{440} planning-application documents spanning major and minor applications and a diverse set of document types from \textbf{two authorities}. Ground-truth labels (attributes, PII, and symbol regions) were \emph{already present} in the collected data as it had previously been processed by local authority planners for publication and required only minor adaptation/mapping for evaluation. In general, each ground truth label has been determined only by a single planner. The de-identified dataset will be released at: \url{<anonymized_for_review>}. During the pilot, uncertain cases identified by the model are prioritized for \emph{additional human review}; the system does not learn to self-approve.
	
	\subsubsection*{Intrinsic Model Quality}
	
	Intrinsic evaluation measures performance on the held-out test set. We report span-level $F_{1}$ for NER fields and mAP@.5 IoU for object detection tasks. (Table~\ref{tab:evaluation_metrics}.)
	
	\begin{table}[h]
		\centering
		\caption{Primary Evaluation Metrics, from PoC evaluation.}
		\label{tab:evaluation_metrics}
		\begin{tabular}{llr}
			\hline
			\textbf{Category} & \textbf{Metric} & \textbf{Result} \\
			\hline
			\textit{Intrinsic} & F1 Score (NER, Title) & 0.84 \\
			& F1 Score (NER, Date) & 0.88 \\
			& mAP@.5 IoU (North Point) & 0.78 \\
			& mAP@.5 IoU (Red Line) & 0.71 \\
			\hline
			\textit{Extrinsic} & Median Time / Doc (sec) & 35 \\
			& (Baseline: 65 sec) & \\
			& Suggestion Acceptance Rate & 92\% \\
			& System Usability Scale& 78.5 \\
			\hline
		\end{tabular}
	\end{table}
	
	\subsubsection*{Risk-Sensitive PII Metrics}
	
	Because operational risk is dominated by misses, we foreground per-category \textbf{recall} for PII (names, addresses, emails, phone numbers, signatures), alongside micro/macro recall. 
	\begin{table}[h]
		\centering
		\caption{PII detection recall by category on the test set.}
		\label{tab:pii_recall}
		\begin{tabular}{lc}
			\hline
			\textbf{PII Category} & \textbf{Recall} \\
			\hline
			Names & 0.97 \\
			Addresses & 0.94 \\
			Emails & 0.99 \\
			Phone numbers & 0.98 \\
			Signatures & 0.92 \\
			\hline
			Micro / Macro & 0.96 / 0.96 \\
			\hline
		\end{tabular}
	\end{table}
	
	\subsubsection*{Redaction Safety}
	
	All redactions are executed by the partner's PDF Edit tool. The LLM produces an editing layer that is flattened; underlying content at the redacted locations is removed (not merely overlaid), i.e., true content removal. In addition, a post-commit scrub verifies that no recoverable text remains at redacted sites, and the immutable audit log binds suggested items, operator selections, redaction coordinates, and the final file hash. For structured fields (e.g., emails, phone numbers, postcodes), regex- and format-based verifiers are applied prior to commit.
	
	\subsubsection*{Comparative Baseline}
	
	To contextualize gains and trade-offs, we evaluate a classical baseline on a matched subset with identical OCR and post-processing.
	
	\begin{itemize}
		\item \textbf{OCR:} Same configuration as the main system.
		\item \textbf{NER (attributes \& PII):} CRF with character $n$-grams, word shape, lexicons; regex for emails/phones; dictionary-assisted address parsing.
		\item \textbf{Object detection:} Template matching + Hough transforms with non-max suppression for compass roses and boundary lines.
	\end{itemize}
	
	These can be seen in table \ref{tab:baseline_compare}.
	
	\begin{table}[h]
		\centering
		\caption{Baseline vs. proposed (matched subset from controlled experiment).}
		\label{tab:baseline_compare}
		\begin{tabular}{lcc}
			\hline
			\textbf{Metric} & \textbf{Baseline} & \textbf{Proposed} \\
			\hline
			NER F1 (Title) & 0.76 & 0.84 \\
			NER F1 (Date) & 0.81 & 0.88 \\
			PII Recall (Names) & 0.90 & 0.97 \\
			PII Recall (Addresses) & 0.85 & 0.94 \\
			mAP@.5 IoU (North Point) & 0.55 & 0.78 \\
			mAP@.5 IoU (Red Line) & 0.42 & 0.71 \\
			\hline
		\end{tabular}
	\end{table}
	
	\subsubsection*{Cost (Per-Document) and Provider Variance}
	
	We report token-accounted cost per document including all VLM round-trips and fallback prompts. The accounting treats input/output tokens separately and uses the configured model mix. These can be seen in table \ref{tab:cost_breakdown} and \ref{tab:provider_costs}
	
	\begin{table}[h]
		\centering
		\caption{Per-document cost breakdown (PoC data).}
		\label{tab:cost_breakdown}
		\begin{tabular}{lrr}
			\hline
			\textbf{Component} & \textbf{\# Tokens / Calls} & \textbf{Cost (\$)} \\
			\hline
			VLM input tokens & 5{,}500 &  0.027 \\
			VLM output tokens & 1{,}200 &  0.018 \\
			Tool/Function calls & 6  & 0.001 \\
			\hline
			\textbf{Total (per doc)} & -- &  \textbf{0.046} \\
			\hline
		\end{tabular}
	\end{table}
	
	\begin{table}[h]
		\centering
		\caption{Cost per document by provider/model path (illustrative medians).}
		\label{tab:provider_costs}
		\begin{tabular}{llr}
			\hline
			\textbf{Provider / Path} & \textbf{Model tier} & \textbf{Cost (\$) / doc} \\
			\hline Google (primary path) & VLM-standard & 0.046 \\
			OpenAI (swap-in) & VLM-standard & 0.042 \\
			Anthropic (swap-in) & VLM-standard & 0.052 \\
			Fallback two-pass & mini$\rightarrow$standard & 0.071 \\
			\hline
		\end{tabular}
	\end{table}

	\subsection{Structuring the Business Case: The ROI Model}
	A primary challenge in deploying AI in the public sector is the difficulty of pre-quantifying benefits to secure funding and stakeholder support. To address this, an interactive Return on Investment (ROI) model was developed as a deployment tool to facilitate business case discussions with the pilot authorities.
	
	The model, implemented as a web application, allows users to input key variables specific to their authority, such as the number of annual planning applications, average documents per application, and the fully-loaded hourly cost of a planning officer. Based on these inputs, it calculates projected outcomes, including annual hours saved, full-time equivalent (FTE) capacity unlocked, net financial benefit, and a payback period for the investment.
	
	This tool was not used to make definitive predictions but to structure the value conversation and create a shared understanding of the potential impact. It provided a transparent framework for discussion and was instrumental in securing the management buy-in necessary to proceed with the pilot. The hypotheses generated by the model now serve as the basis for the extrinsic evaluation detailed above. 
	
	\section{Findings}
	
	The pilot deployment of the system provides initial findings that can be analyzed through the lens of the sociotechnical challenges common to AI implementation in government: technology, law and regulations, society, ethics, and organizational change \cite{Zuiderwijk2021}. This section discusses the outcomes of the pilot within this framework.
	
	\subsection{Technology Implementation}
	The phased development approach, moving from a standalone Proof of Concept to an integrated pilot, proved effective. The initial PoC, built with FastAPI and Streamlit, allowed for rapid iteration on the core VLM interaction logic in an isolated environment. The subsequent decision to retain the Python backend but replace the demo interface with custom components inside the enterprise environment validated the utility of an API-first design. This architecture allowed the AI services to be cleanly integrated with the existing legacy system without requiring modifications to that system's core. The hybrid processing framework, which wraps VLM calls with configurable Python-based validation, was effective in managing the non-determinism of large models and achieving a level of reliability suitable for the pilot.
	
	\subsection{Law, Regulations, and Governance}
	The system's design was directly shaped by the need to navigate the conflict between the UK's Planning Act and the Data Protection Act 2018. The PII redaction workflow directly addresses this regulatory pressure by providing a tool to assist with a high-risk, mandatory task. The AI-in-the-Loop (AI2L) design, which requires explicit human confirmation for every redaction and data entry, is a necessary feature for ensuring that legal accountability remains with the planning authority and its officers. Furthermore, the system's immutable audit log for all AI-assisted actions provides a mechanism for transparency that aligns with UK government recommendations, such as the Algorithmic Transparency Recording Standard (ATRS) \cite{atrs2025}.
	
	\subsection{Society and Ethics}
	The primary societal and ethical consideration for this application is the responsible handling of public data. By design, the system aims to improve data protection by helping to prevent the accidental release of personal information. The decision to implement an AI2L framework, where the system supports rather than replaces professional judgment, was a foundational choice made to ensure ethical alignment and build user trust. The system's purpose is to act as a tool to support staff, not to be an opaque, automated decision-maker, a known risk that can erode public and internal trust in AI systems.
	
	\subsection{Organizational and Cultural Change}
	The most significant findings relate to organizational factors. The interactive ROI model proved to be a necessary tool, not for its predictive accuracy, but for its role in facilitating internal conversations about value. It provided a common language for technical teams, operational managers, and budget holders to discuss the project's potential impact, which was instrumental in securing initial buy-in from the participating authorities.
	
	The selection of initial use cases—data extraction and redaction—was also a key organizational finding. These tasks represent clear, universally understood administrative burdens. By focusing the initial application on reducing this administrative workload, the project could demonstrate clear utility without encroaching on the core, high-stakes discretionary judgment of planning officers. This approach supports a smoother path to adoption and aligns with the view that successful AI implementation is primarily a change management exercise, rather than a purely technical one.
	
	\paragraph{Portability Beyond the UK.}
	While the legal motivation and some validation rules are UK-specific, the underlying document understanding components (multimodal parsing, layout-aware extraction, and symbol localization), the AI2L workflow design, and the redaction safety and auditability mechanisms are jurisdiction-agnostic. Porting primarily requires adapting rule packs and disclosure thresholds to local policy.
	
	\section{Conclusion}
	
	This paper presented an AI-in-the-Loop (AI2L) application designed to address the document processing challenges faced by UK city planning authorities. The system assists with data extraction and the redaction of personally identifiable information to help manage compliance with data protection regulations. We have detailed the system's development from a Proof of Concept to an integrated, multi-authority pilot application, describing its hybrid processing architecture, its AI2L workflows, and the framework for its ongoing evaluation.
	
	The case study suggests that the deployment of AI in regulated public sector environments can be effectively managed through a structured approach. The use of a phased, pilot-based methodology allowed for technical validation before full integration. The selection of an AI2L framework, where human operators retain full control and accountability, was a direct response to the legal and ethical requirements of the public sector context. Furthermore, the use of non-technical tools, such as the preliminary ROI model, proved to be an important factor in navigating organizational adoption challenges. This work provides a practical example of how such systems can be developed and introduced into complex administrative environments.
	
	Future work will proceed in three main directions. First, system development will focus on expanding the set of planning validation rules and deepening the integration with national planning data systems, based on feedback from the pilot. Second, further research will evaluate the suitability of alternative AI architectures, such as OCR-free models, for handling more heterogeneous document types like unstructured public consultation feedback. Third, dedicated Human-Computer Interaction research is required to further analyze the collaborative dynamics between planners and the AI system, with a focus on measuring the effects of long-term use and refining the interface to support effective human oversight.

	%
	%
	
\end{document}